\documentclass[conference]{IEEEtran}
\IEEEoverridecommandlockouts
\usepackage{amsmath,amssymb,amsfonts}
\usepackage{algorithmic}
\usepackage{textcomp}
\usepackage{xcolor}

\usepackage{url}
\usepackage{graphicx}
\usepackage{amsmath}
\usepackage{amsthm}
\usepackage{dsfont}
\usepackage{subcaption}
\usepackage{color}
\usepackage{enumerate}
\usepackage{cases}
\usepackage{multicol}
\usepackage{multirow}
\usepackage{enumitem}
\usepackage{bm}
\usepackage{import}
\usepackage{lscape}
\usepackage{xcolor}
\usepackage{float}
\usepackage{booktabs}       

\usepackage[algo2e,linesnumbered,ruled,vlined]{algorithm2e}
\usepackage{caption}
\usepackage{wrapfig}

\definecolor{BrightBlue}{RGB}{65, 145, 225}

\definecolor{figure_blue}{RGB}{53, 132, 187}
\definecolor{figure_orange}{RGB}{255, 139, 38}
\definecolor{figure_green}{RGB}{65, 169, 65}
\definecolor{figure_red}{RGB}{218, 60, 61}
\definecolor{figure_purple}{RGB}{158, 118, 195}
\definecolor{figure_brown}{RGB}{160, 82, 45}
\definecolor{figure_pink}{RGB}{255,105,180}

\makeatletter
\newcommand\footnoteref[1]{\protected@xdef\@thefnmark{\ref{#1}}\@footnotemark}
\makeatother

\def\BibTeX{{\rm B\kern-.05em{\sc i\kern-.025em b}\kern-.08em
    T\kern-.1667em\lower.7ex\hbox{E}\kern-.125emX}}
\begin{document}

\title{Lazy But Effective: Collaborative Personalized Federated
Learning with Heterogeneous Data}


\author{\IEEEauthorblockN{1\textsuperscript{st} Ljubomir Rokvic}
\IEEEauthorblockA{\textit{Artificial Intelligence Laboratory} \\
\textit{EPFL}\\
Lausanne, Switzerland \\
ljubomir.rokvic@epfl.ch}
\and
\IEEEauthorblockN{2\textsuperscript{nd} Panayiotis Danassis}
\IEEEauthorblockA{\textit{Telenor Research} \\
Oslo, Norway \\
panayiotis.danassis@telenor.com}
\and
\IEEEauthorblockN{3\textsuperscript{rd} Boi Faltings}
\IEEEauthorblockA{\textit{Artificial Intelligence Laboratory} \\
\textit{EPFL}\\
Lausanne, Switzerland \\
boi.faltings@epfl.ch}
}

\maketitle

\begin{abstract}
In Federated Learning, heterogeneity in client data distributions often means that a single global model does not have the best performance for individual clients. Consider for example training a next-word prediction model for keyboards: user-specific language patterns due to demographics (dialect, age, etc.), language proficiency, and writing style result in a highly non-IID dataset across clients. Other examples are medical images taken with different machines, or driving data from different vehicle types. To address this, we propose a simple yet effective personalized federated learning framework (\emph{pFedLIA}) that utilizes a computationally efficient influence approximation, called \emph{`Lazy Influence'}, to cluster clients in a distributed manner before model aggregation. Within each cluster, data owners collaborate to jointly train a model that captures the specific data patterns of the clients. Our method has been shown to successfully recover the global model's performance drop due to the non-IID-ness in various synthetic and \emph{real-world} settings, specifically a next-word prediction task on the Nordic languages as well as several benchmark tasks. \emph{It matches the performance of a hypothetical Oracle clustering, and significantly improves on existing baselines, e.g., an improvement of 17\% on CIFAR100.}
\end{abstract}


\section{Introduction} \label{sec:Introduction}

Federated Learning (FL) \cite{mcmahan2017communication,kairouz2021advances,wang2021field,li2020federated} has emerged as a promising paradigm for distributed learning across a multitude of decentralized devices, such as mobile phones, edge devices, sensors etc. In FL, a central server coordinates the training of a global model across multiple decentralized clients, each holding its own local data that remains on-device. This approach enables collaborative model training without sharing raw data, thereby preserving privacy and reducing communication overhead. In this paper, we propose an efficient algorithm that addresses a central challenge in FL: handling the inherent \emph{heterogeneity} in client data distributions.

In real-world scenarios, data across clients are often highly non-IID, leading to a global model that may perform suboptimally for individual clients. For instance, users of a virtual keyboard exhibit unique language patterns influenced by a plethora of factors such as dialect, age, writing style, etc. Training a single global model for next-word prediction without accounting for this diversity can result in poor personalization and user experience (e.g., having a model that uses the most popular dialect, instead of the user's). Other examples include personalized advertisements according to different groups of customers, medical image segmentation captured by different devices, driving models for different vehicle types, etc.

\begin{figure}
    \centering
    \includegraphics[width=\linewidth, clip, trim={0em 0cm 0em 0em}]{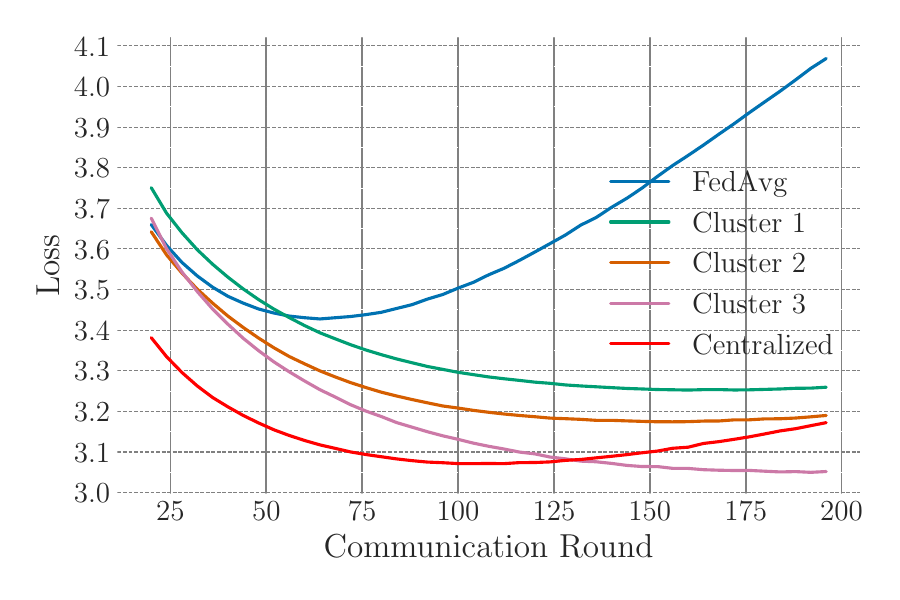}
    \caption{Validation loss over time, when training a GPT-2 model for next word prediction. We compare using a single model trained on all client data using FedAvg (blue line), versus clustering clients in a peer-to-peer setting using the proposed pFedLIA and learning distinct models. pFedLIA automatically splits the clients into three clusters (green, orange, and pink lines, respectively -- which roughly correspond to Norwegian, Swedish, and Danish clients). FedAvg is clearly unable to learn in this setting, due to the \emph{data heterogeneity}. In comparison, pFedLIA achieves loss comparable to centralized training, where all data is collected, and a single model is trained centrally (red line).}
    \label{figure: NLP}
\end{figure}

\begin{figure}
    \centering
    \includegraphics[width=\linewidth, clip, trim={0em 18cm 0em 0em}]{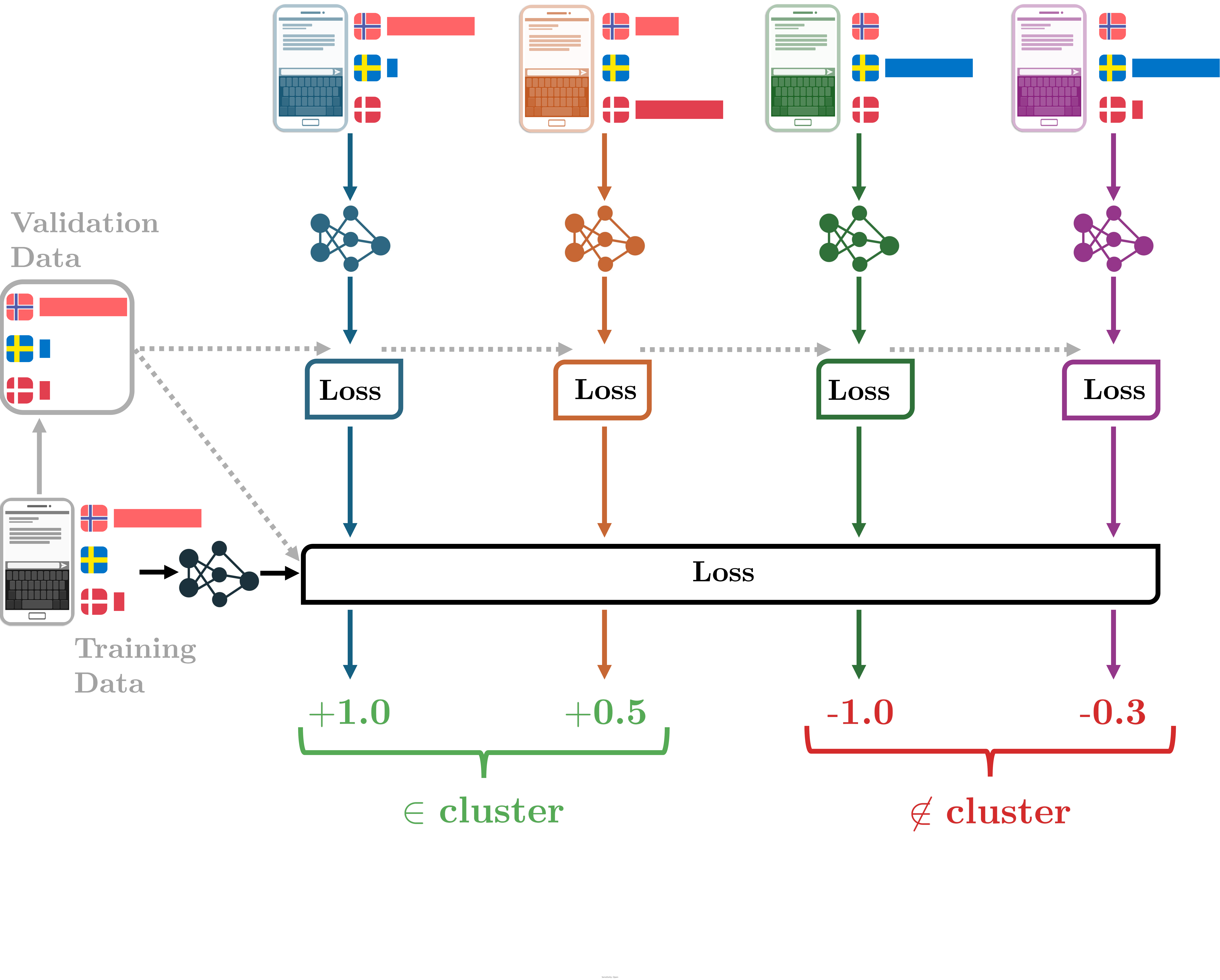}
    \caption{Visual illustration of the proposed pFedLIA, from the perspective of client $i$ (black phone on the left). Client $i$ calculates a \emph{Lazy Influence Approximation} score for every other client $j$ (phones on the top). To do so, every client $j$ performs a few local epochs to the joint model $M_0$ -- enough to get an estimate of the gradient (e.g., 5–20), no need to train the model fully -- using a batch of their training data. Subsequently, $j$ sends the updated partial model $M_{j}$ to client $i$. The Lazy Influence Approximation $\mathcal{I}(i,j)$ is simply the difference in the validation loss of models $M_0$ and $M_{j}$. Finally, we use these influence scores to cluster clients; in this case, the blue and orange phones will be in $i$'s cluster, while the green and purple will not.}
    \label{figure: pFedLIA Illustration}
\end{figure}

A common theme in all of these applications is that clients can naturally be partitioned into 
clusters \cite{ghosh2020efficient,mansour2020three}; e.g., users speaking different dialects, or interested in different categories of news. Then, one can tailor models to individual clusters of clients with similar data distributions by learning \emph{personalized models} (Personalized FL~\cite{zhang2020personalized}). Naturally, the main challenge in this setting is determining the cluster of each client. Traditionally, clients in clustering-based Personalized FL (PFL) approaches assume a passive role (data suppliers). Instead, we focus on a \emph{client-centric} PFL setting, allowing each client to \emph{choose the subset of ‘relevant’ clients} to aggregate models, based on client-specific downstream tasks.

In this paper, we propose a simple yet effective clustering federated learning framework, \emph{pFedLIA} (personalised Federated learning using Lazy Influence Approximation), that leverages a computationally efficient influence approximation, adapted from the \emph{Lazy Influence} of~\cite{Rokvic2024LIA}, to cluster clients prior to model aggregation. Using the \emph{locally computed} influence scores, clustering can then be performed either centrally, or individually by every client in a \emph{peer-to-peer} setting. The latter offers two significant advantages: (i) it alleviates the computational cost at the center, and (ii) allows for fine-grained personalized models for \emph{decentralized clustering}, where each client can aggregate data from any other client.

As a motivating example, we trained a GPT-2 model for next word prediction. We chose the Scandinavian languages (Norwegian, Swedish, and Danish), as they are closely related, and largely mutually intelligible, particularly in their standard varieties. Speakers of these three languages can broadly read all three without great difficulty~\cite{ed7ea7ee31534584be17e2dcd4e7ba7}. We used \emph{real-data} for training, from the OSCAR corpus~\cite{ortiz-suarez-etal-2020-monolingual}. Figure~\ref{figure: NLP} shows the validation loss over time, if we use a single model trained on all client data using FedAvg (blue line), compared to clustering clients in a peer-to-peer setting using pFedLIA and learning distinct models. It is important to note that pFedLIA \emph{automatically splits the clients} into the 3 relevant clusters -- \emph{it does not require any a priori knowledge of the number of clusters}. FedAvg is clearly unable to learn in this setting, due to the data heterogeneity, while the proposed approach, pFedLIA, is able to successfully learn and achieve loss close to the centralized model, where all data are collected and a single model is trained centrally (red line).

\subsection{High-Level Description of Our Approach} \label{sec: High-Level Description of Our Approach}

In traditional Federated Learning, clients assume the passive role of data suppliers. As a result, all clients receive the same global model. Instead, in this paper, we focus on a \emph{client-centric} personalized Federated Learning setting, where each client adopts an \emph{active} role of choosing a subset of `relevant' clients to aggregate models, based on client-specific downstream tasks and unique data distributions. The result is a \emph{set} of models -- one per cluster of clients.

Clustering clients can be performed in a centralized or decentralized manner. In both cases, the clients must first compute the influence of their peers in their own unique distribution. Subsequently, in the centralized clustering case, a center $C$ collects all the influence scores -- forming an $N \times N$ array, where $N$ is the number of clients -- and uses any off-the-self density-based clustering algorithm (e.g., OPTICS~\cite{ankerst1999optics}) to form client clusters. Note that the OPTICS algorithm does not require knowing the number of clusters in advance. In the decentralized clustering case, each client forms a vector (of size $N$) of influence scores, and uses k-means to form two clusters (representing the inclusion or exclusion of clients from the aggregation).

The most computationally expensive part of the aforementioned process is computing the influence scores. In simple terms, influence quantifies the marginal contribution of a data point (or, in our case, batch of points) on a model's accuracy. A positive influence value indicates that a data point improves model accuracy, and vice versa. One can compute this by comparing the difference in the model's empirical risk when trained with and without the point in question. While the influence metric can be highly informative, it is impractical to compute: re-training a model is time-consuming, costly, and often impossible, as clients do not have access to other clients' data. Instead, we adapt a simple and practical approximation of the exact influence (\emph{Lazy Influence}~\cite{Rokvic2024LIA}) which is based on an estimate of the direction of the model after a very small number (e.g., 10–20) of local training epochs 
with the new data. 

Concretely, suppose that client $i$ wants to evaluate the `benefit' (influence score) of incorporating the data of client $j$ in their cluster. We assume that both clients have access to an initial model $M_0$ that captures the desired input/output relation. This can be trained, e.g., by the center $C$ using a small set of `warm-up' data. Client $j$ performs a few local epochs to $M_0$ -- enough to get an estimate of the gradient (e.g., 10–20), no need to train the model fully -- using a batch of their training data $z_{j}$. Subsequently, $j$ sends the updated partial model $M_{j}$ to client $i$. The Lazy Influence Approximation (as adapted for our setting) $\mathcal{I}(i,j)$ is simply the difference in the validation loss of models $M_0$ and $M_{j}$. Note that if one wanted to compute the actual influence, then client $j$ would require access to all the data used to train $M_0$ (denoted as $D_0$), and then proceed to fully train a model until `convergence' using all data $D_0 \cup z_{j}$. This is impractical in real setting due to both computation limitations, but, importantly, lack of access to training data.

In summary, the proposed \emph{pFedLIA} works as follows (see Figure \ref{figure: pFedLIA Illustration}). Every client $i$ calculates a score for every other client $j$, based on an adaptation of the \emph{Lazy Influence Approximation}. In the case of centralized clustering, the center collects these values and clusters the clients (e.g., using OPTICS). Importantly, \emph{the center does not need to know the number of clusters in advance}. Alternatively, each client can perform their own clustering using k-means. Finally, the models in each cluster are then updated with the contributions of the participants using \emph{any} aggregation method for federated learning (e.g., FedAvg). Note that the clustering process is performed only \emph{once} by every client, when they enter the federation.

\subsection{Our Contributions} \label{sec: Our Contributions}

\begin{itemize}
\item We present a novel collaborative personalized Federated Learning algorithm, \textbf{pFedLIA} (personalized Federated learning using Lazy Influence Approximation), that leverages a lazy influence score, to cluster clients prior to model aggregation. Clustering in pFedLIA can be performed both centrally by the center of the federation, but also decentralized by each client in a \textbf{peer-to-peer} setting. The latter allows each client to \textbf{choose their own subset of ‘relevant’ clients}. Clustering is only performed \textbf{once}, and \textbf{pFedLIA does not require knowing the number of clusters}. pFedLIA can be combined with any aggregation technique for federated learning.
\item We show that pFedLIA is able to successfully \textbf{recover from the model's performance drop due to the data heterogeneity} in various synthetic and \emph{real-world} settings, specifically a next-word prediction task on the Nordic languages as well as several well-known benchmark tasks. \textbf{It matches the performance of a hypothetical Oracle clustering, and outperforms the best baseline by 17\% on CIFAR100.}
\end{itemize}

\section{Related Work and Discussion} \label{sec: Related Work}


\emph{(a) Federated Learning (FL)}~\cite{mcmahan2017communication,kairouz2021advances,wang2021field,li2020federated} has emerged as an alternative method to train ML models on data obtained by many clients. In FL, a center coordinates clients who acquire data and provide model updates. FL has been receiving increasing attention in both academia~\cite{lim2020federated,yang2019federated,he2020fedml,caldas2018leaf} and industry~\cite{hard2018federated,chen2019federated}, with a plethora of real-world applications such as training models from smartphone data, IoT devices, sensors, etc. 

The intuitive strategy, of fine-tuning a centralized global model on an client's personal data, can lead to a plethora of issues such as parameter divergence~\cite{DBLP:journals/corr/abs-1806-00582}, failure to converge ~\cite{li2020federated}, and various biases~\cite{hsieh2019non}. These problems are a common side effect of data heterogeneity in decentralized clients. In recent years, there have been substantial improvements to the robustness of global models under skewed, non-IID settings. One approach
is to introduce momentum to the global aggregation, as defined in \textit{FedAvgM}~\cite{hsu2019measuring}. \textit{SCAFFOLD} is a method proposed in~\cite{karimireddy2020scaffold}, which reduces the client-drift by using variance reduction. \cite{zhu2021data} propose for the server to learn a lightweight generator to ensemble user information in a data-free manner, to tackle the problem of data heterogeneity.

These methods may improve the robustness of global model aggregation across data heterogeneous settings, but they do not address the per client performance on their respective local data. While the overall global performance might improve by a noticeable amount, the per client on device performance should be the priority. Their local data distributions may not match the global model, resulting in drastically reduced performance. Importantly, our proposed approach can also be combined with and benefit by robust aggregation techniques, to further improve performance. Yet, this is out of the scope of this work.


\emph{(b) Personalized Federated Learning (PFL)} attempts to tackle the aforementioned challenges in a variety of ways~\cite{tan2023pfedsim, li2019fedmd, li2021fedbn, arivazhagan2019federated, husnoo2022fedrep, t2020personalized,  shamsian2021personalized, ma2022layer, oh2021fedbabu, yeganeh2022fedap, chen2023metafed, chen2021bridging, scott2024pefll, deng2020adaptive, hanzely2020federated, mansour2020three,zhang2020personalized, guo2022adaptive}. One proposed way to tackle these issues is to train multiple models to match the target distributions. Such a method is proposed by~\cite{DBLP:journals/corr/SmithCST17}, where the challenge is presented as a multitask learning problem in which one model corresponds to one client. Another proposed approach is the use of global semantic knowledge for learning better representations~\cite{xu2023personalized}.

The personalized FL that is most relevant to our work is Clustered FL~\cite{long2023multi, sattler2020clustered, li2021federated, kim2021dynamic, ghosh2020efficient, briggs2020federated, duan2021flexible, ouyang2022clusterfl, mansour2020three, tang2024fuzzy}. For a comprehensive overview of Personalized Federated Learning, we direct the interested reader to~\cite{tan2022towards}. Most of these methods study a Non-IID setting created by augmenting the data (e.g., rotating CIFAR10 images\cite{ghosh2020efficient}) or by sampling different devices~\cite{ouyang2022clusterfl}. While this feature space heterogeneity is an important problem, our main focus will be label space heterogeneity (though we demonstrate that our method works for both). The settings described in~\cite{long2023multi, sattler2020clustered} are of much more interest for our method. We directly compare to these well established gradient similarity based methods.


\emph{(c) Influence functions} are a standard method from robust statistics~\cite{cook1980characterizations} (see also Section \ref{sec: Methodology}), which were recently used as a method of explaining the predictions of black-box models \cite{pmlrv70koh17a}. They have also been used in the context of fast cross-validation in kernel methods and model robustness \cite{liu2014efficient,christmann2004robustness}. While a powerful tool, computing the influence involves too much computation and communication, and it requires access to the training and validation data (see~\cite{pmlrv70koh17a} and Section \ref{sec: Methodology}). There has also been recent work trying to combine Federated Learning with influence functions~\cite{xue2021toward}.


\emph{(d) Distributed Learning} has a lot of similarities with the area of Federated Learning, with the key difference being that there exists no central authority, but all the learning is done between the clients~\cite{verbraeken2020survey}. The main benefits of this approach are a focus on individual client performance and flexibility in learning. Our proposed method is naturally applicable in peer-to-peer learning (e.g., see~\cite{bellet2018personalized}), as both the Lazy Influence calculation and the subsequent clustering can be performed on a per-client basis as we describe in Section \ref{sec: Methodology}.

\section{Methodology}  \label{sec: Methodology}

This work naturally applies to heterogeneous settings, where clients' data are not necessarily sampled from a common distribution, and we have \emph{no a priori knowledge} of the number or characteristics of these distributions (and subsequently the downstream tasks). In this section, we focus a supervised classification problem as our setting, but the proposed approach is applicable to any task with a well-defined loss function (in fact, we have evaluated the proposed approach in a next-word prediction task).

We aim to address two challenges: (i) approximating the influence of a (batch of) data point(s) without having to re-train the entire model from scratch and (ii) leverage this signal to distinguish the underlying local distributions and effectively cluster clients without sharing any data. We first introduce the notion of \emph{influence}~\cite{cook1980characterizations} and the adapted (from~\cite{Rokvic2024LIA}) lazy approximation. Then, we describe the proposed pFedLIA (personalized Federated learning using Lazy Influence Approximation) algorithm, an client-centric FL framework to efficiently learn effective combinations of the available models for each participating client.

\subsection{Setting}  \label{sec: Setting}

Let $\mathcal{N}$ denote the set of clients. We consider a classification problem from some input space $\mathcal{X}$ (e.g., features, images, etc.) to an output space $\mathcal{Y}$ (e.g., labels). We want to learn a model $M(\theta)$ parameterized by $\theta \in \Theta$, with a non-negative loss function $L(z, \theta)$ on a sample $z=(\bar{x},y) \in \mathcal{X} \times \mathcal{Y}$. Let $R(Z, \theta) = \frac{1}{n}\sum_{i=1}^n L(z_i, \theta)$ denote the empirical risk, given a set of data $Z = \{z_i\}_{i=1}^n$. We assume that the empirical risk is differentiable in $\theta$. 

\subsection{Exact Influence Definition}

In simple terms, the influence measures the marginal contribution of a data point on a model's accuracy. A positive influence value indicates that a data point improves model accuracy, and vice versa. More specifically, let $Z = \{z_i\}_{i=1}^n$,  $Z_{+j} = Z \cup z_j$ where $z_j \not\in Z$, and let $\hat{R} = \min_{\theta} R(Z, \theta)$ and $\hat{R}_{+j} = \min_{\theta} R(Z_{+j}, \theta)$, where $\hat{R}$ and $\hat{R}_{+j}$ denote the minimum empirical risk of their respective set of data. The \emph{influence} of datapoint $z_j$ on $Z$ is defined as $\mathcal{I}(z_j,Z) \triangleq \hat{R} - \hat{R}_{+j}$.

\subsection{Shortcomings of the Exact and Approximate Influence in an FL Setting}  \label{sec: Influence}

Despite being highly informative, influence functions have not achieved widespread use in FL (or ML in general). This is mainly due to the computational cost. The exact influence requires complete retraining of the model, which is time-consuming and very costly, especially for state-of-the-art, large ML models (importantly for our setting, we do not have direct access to the training data). Recently, the first-order Taylor approximation of influence~\cite{pmlrv70koh17a} (based on~\cite{cook1982residuals}) has been proposed as a practical method to understanding the effects of training points on the predictions of a \emph{centralized} ML model. While it can be computed without having to re-train the model, according to the following equation $\mathcal{I}_{appr}(z_{j}, z_{val}) \triangleq - \nabla_\theta L(z_{val}, \hat{\theta}) H^{-1}_{\hat{\theta}} \nabla_\theta L(z_j, \hat{\theta})$, it is still ill-matched for FL models for several key reasons.

Computing the influence approximation of~\cite{pmlrv70koh17a} requires \emph{forming and inverting} the Hessian of the empirical risk. With $n$ training points and $\theta \in \mathbb{R}^m$, this requires $O(nm^2 + m^3)$ operations~\cite{pmlrv70koh17a}, which is \emph{impractical} for modern-day deep neural networks with millions of parameters. To overcome these challenges,~\cite{pmlrv70koh17a} used implicit Hessian-vector products (iHVPs) to more efficiently approximate $\nabla_\theta L(z_{val}, \hat{\theta}) H^{-1}_{\hat{\theta}}$, which is typically $O(m)$. While this is a somewhat more efficient computation, it is \emph{communication-intensive}, as it requires \emph{transferring all the data} at each FL round. 
Finally, the loss function has to be strictly convex and twice differentiable (which is not always the case in modern ML applications). \cite{pmlrv70koh17a} proposed to swap out non-differentiable components for smoothed approximations, but there is no quality guarantee of the influence calculated in this way.

\subsection{Lazy Influence Approximation (LIA)} \label{sec: LIA}


The key realization is that \emph{we do not need to compute an exact influence value to effectively cluster clients}; we only need an `accurate enough' estimate of its direction. Recall that a positive influence value indicates a data point improves model accuracy. We employ a \emph{fast, stripped-down adaptation} of the `Lazy Influence Approximation' of~\cite{Rokvic2024LIA}. Calculating the proposed `Lazy Influence' works as follows (see also Figure~\ref{figure: pFedLIA Illustration}). Let $\theta_0$ denote an initial model that captures the desired input/output relation. This can be trained, e.g., by the center $C$ using a small set of `warm-up' data. 

\begin{enumerate}
    \item Every client $j \in \mathcal{N}$ performs a small number $k$ of local epochs to $\theta_0$ using a batch of his training data $Z_{j}$, resulting in $\tilde{\theta}_j$. $k$ is a hyperparameter. $\tilde{\theta}_j$ is the partially trained model of participant $j$. The model should \emph{not} be fully trained for two key reasons: efficiency and avoiding over-fitting (in our simulations, we only performed 20 epochs). Finally, $j$ sends $\tilde{\theta}_j$ to every other participant.
    \item Participant $i$, upon receiving $\tilde{\theta}_j$, uses his validation dataset $Z^{val}_i$ to estimate the influence as: 
    \begin{equation} \label{eq: proposed influence}
        \mathcal{I}_{LIA}(i, j) \triangleq \sum_{z \in Z^{val}_i}  L(z, \theta_0)  -  L(z, \theta_j)
    \end{equation}
\end{enumerate}

Finally, either the center $C$ aggregates the influence values and performs clustering, or, in a peer-to-peer setting, each client does so locally, as we explain in Section~\ref{sec: pFedLIA}.


\begin{figure}[t]
	\centering
	\begin{subfigure}{.45\linewidth}
		\centering
		\includegraphics[width=\linewidth, clip, trim={0em 1.5em 0em 4em}]{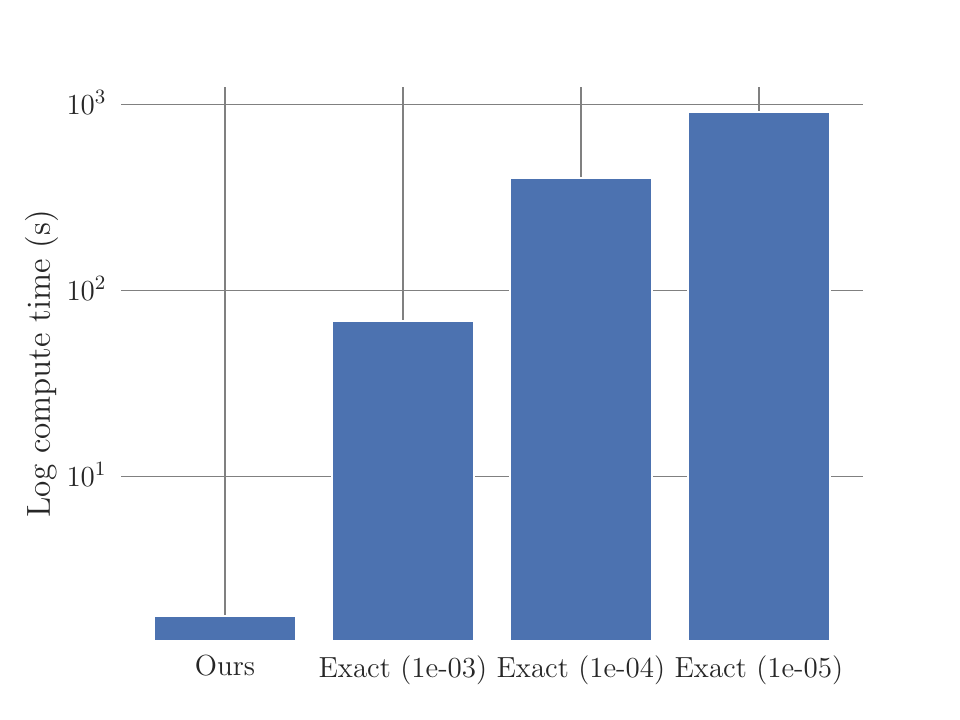}
		\caption{}
		\label{figure: LIA speedup}
	\end{subfigure}%
	\begin{subfigure}{.55\linewidth}
		\centering
		\includegraphics[width=0.8\linewidth, clip, trim={3em 11.5em 40em 10.5em}]{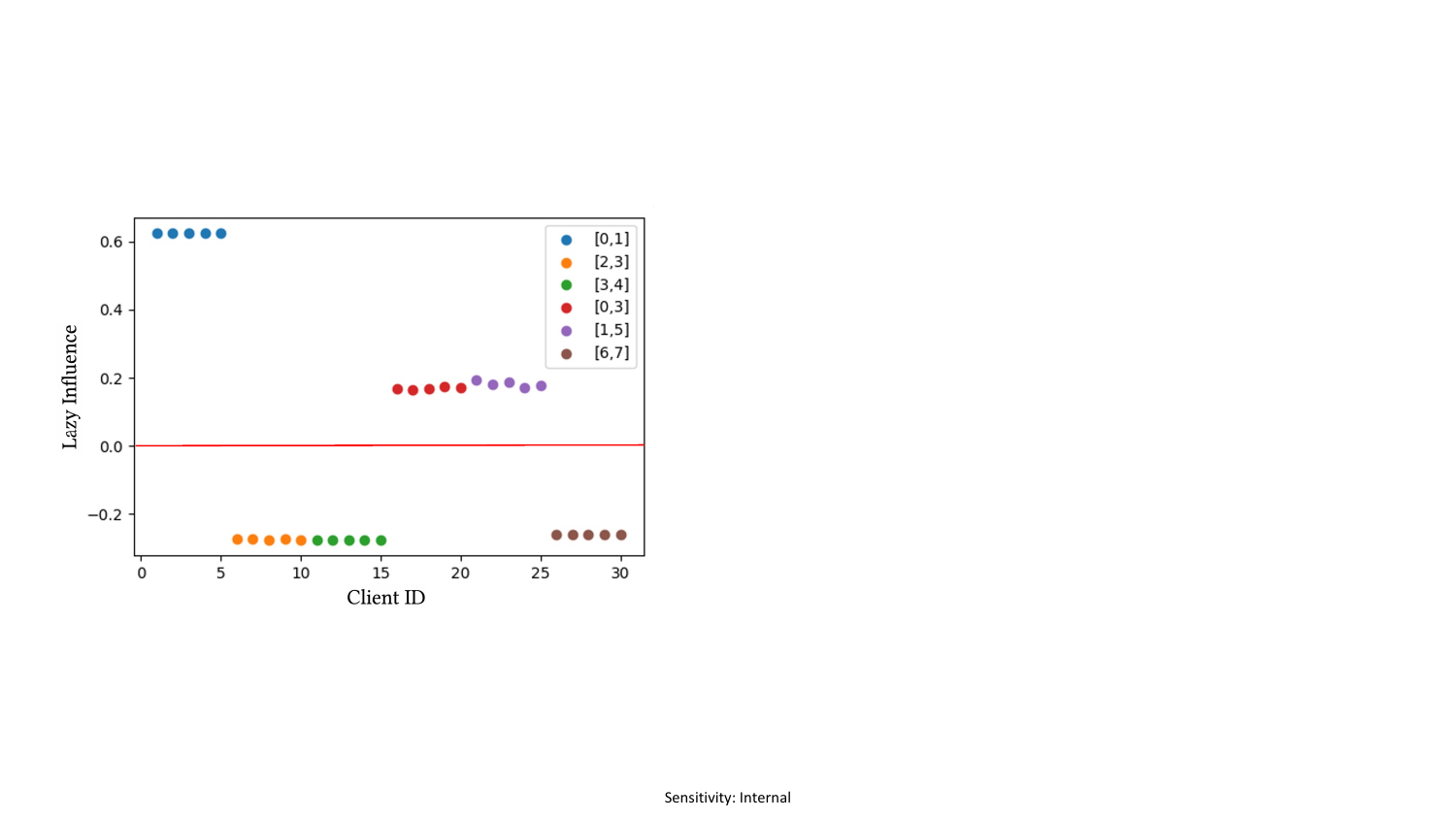}
		\caption{}
		\label{figure: pFedLIA decentralized clustering illustration}
	\end{subfigure}
	\caption{\textbf{(\ref{figure: LIA speedup})} Computation time comparison between the proposed Lazy Influence Approximation (LIA), versus the exact influence (for three different convergence thresholds of varying precision). LIA achieves between \emph{40–500 times speedup}. \textbf{(\ref{figure: pFedLIA decentralized clustering illustration})} pFedLIA decentralized clustering illustration. Consider a classification task (classes [0-7]). We have 32 clients, each assigned 2 out of the 8 classes. This figure depicts the Lazy Influence Approximation values of every client, as calculated by client 0 with data from classes [0, 1]. The red horizontal line represents the clustering frontier, using k-means (2 classes, representing beneficial or not beneficial clients for client 0).}
\end{figure}

\subsection{Advantages of The Employed Lazy Influence}

The designer may select \emph{any optimizer} to perform the model updates. We do not require the loss function to be twice differentiable and convex; only once differentiable. It is significantly more \emph{computation and communication efficient}; an essential prerequisite for any FL application. This is because client $j$ does not need to send their training data. Moreover, computing a few (10-20) model updates (using, e.g., SGD) is significantly faster than computing either the exact influence or an approximation due to the numerous challenges.

As a concrete example, we plot the computation time of the employed adaptation of LIA, compared to the exact influence (Figure~\ref{figure: LIA speedup}). \emph{LIA achieves between 40–500 times speedup}. 

We computed the exact influence as follows: We train a model $M_0$ on an initial dataset $D_0$, which has 9900 datapoints. We wish to evaluate the influence of a batch $B$ (100 data points) in respect to $M_0$. To do so, we initialize a new model $M_1$ and train it on $D_1 = D_0 \cup B$, until we reach a certain convergence threshold (of course, a model can not fully converge, so we need to pick a threshold). Finally, we evaluate both models $M_0$ and $M_1$ on an IID testset (100 data points); the difference in loss is the exact influence metric. We measure the computation time as the training and evaluation of $M_1$ (i.e., we do not include the training time for $M_0$). It is important to note that we opted for this approach, instead of e.g., the approximate influence of~\cite{pmlrv70koh17a}, because calculating approximate iHVPs is impractical for large scale modern ML models (we used Res-Net18~\cite{he2016deep}, which has $\sim12$ million parameters, and we faced both time and memory complexity issues in our local machines).


\subsection{pFedLIA: personalized Federated learning using Lazy Influence Approximation} \label{sec: pFedLIA}

The proposed \emph{pFedLIA} works as follows (see also Figure \ref{figure: pFedLIA Illustration}). Every client $i$ calculates the proposed \emph{Lazy Influence Approximation} for every other client $j$, as described in Section~\ref{sec: LIA}. Then, clustering is performed in a centralized or decentralized, peer-to-peer manner. Finally, clients may use \emph{any} FL aggregator (e.g., FedAvg) to train one model per cluster, resulting in models that better align with each client's downstream task and dataset, as compared to giving all clients the same global model. Note that the clustering process is performed only \emph{once}, contrary to other clustering-based personalized FL approaches (e.g., \cite{long2023multi, sattler2020clustered}). 

\paragraph{Centralized}

In the case of centralized clustering, the center collects these values and clusters the clients using OPTICS~\cite{ankerst1999optics}, a density based clustering algorithm that \emph{does not require knowledge of the number of clusters}. 

\paragraph{Peer-to-Peer}

Alternatively, in the case of decentralized clustering, each client can run k-means. The number of clusters for k-means is always two, representing beneficial or not beneficial clients. As a simple, illustrative example, Figure~\ref{figure: pFedLIA decentralized clustering illustration} shows the LIA values for a classification task, from the perspective of an client in the decentralized setting. The red horizontal line represents the clustering frontier.

Peer-to-peer clustering happens of course at the expense of communication complexity, but offers two significant advantages: (i) it alleviates the computational cost at the center, and (ii) allows for fine-grained personalized models for decentralized clustering, where each client can aggregate data from any other client. Importantly, there are settings where distributed learning is the only viable option.

\section{Evaluation} \label{sec: Evaluation}

\begin{table*}[ht]
\centering
\caption{Results on accuracy compared to SOTA baselines. The proposed approach outperforms all baselines in all (but one) settings (by up to 17\%), and importantly, achieves similar performance to the hypothetical Oracle.}
\label{tb: Results}

\begin{tabular}{l|ccc|ccc}
\toprule
& \multicolumn{3}{c|}{\textbf{Pathological Non-IID Setting}} & \multicolumn{3}{c}{\textbf{Noisy Non-IID setting}} \\
\textbf{Method} & \textbf{CIFAR10} & \textbf{Fashion MNIST} & \textbf{CIFAR100} & \textbf{CIFAR10} & \textbf{Fashion MNIST} & \textbf{CIFAR100} \\
\midrule
Oracle & $\textbf{86.08\%} \pm 1.67\%$ & $\textbf{99.07\%} \pm 0.21\%$ & $\textbf{68.28\%} \pm 2.24\%$ & -- & -- & -- \\
\midrule
Local-only & $72.83\% \pm 4.09\%$ & $84.71\% \pm 5.47\%$ & $34.39\% \pm 1.40\%$ & $66.24\% \pm 2.59\%$ & $84.71\% \pm 5.47\%$ & $33.89\% \pm 1.51\%$ \\
\midrule
FedAvg & $24.43\% \pm 4.84\%$ & $60.99\% \pm 5.07\%$ & $48.20\% \pm 3.40\%$ & $29.09\% \pm 5.03\%$ & $65.95\% \pm 5.37\%$ & $24.52\% \pm 20.67\%$ \\
FedFomo & $80.98\% \pm 2.45\%$ & $98.63\% \pm 0.44\%$ & $43.77\% \pm 4.16\%$ & $69.98\% \pm 1.12\%$ & $ 85.11\% \pm 3.41\%$ & $40.06\% \pm 6.08\%$ \\
FedProto & $67.70\% \pm 6.36\%$ & $93.93\% \pm 1.42\%$ & $36.68\% \pm 2.49\%$ & $65.45\% \pm 2.11\%$ & $\textbf{87.39\%} \pm 2.87\%$ & $33.35\% \pm 2.36\%$ \\
FeSEM & $72.90\% \pm 8.24\%$ & $97.97\% \pm 0.46\%$ & $51.20\% \pm 4.90\%$ & $68.86\% \pm 4.29\%$ & $81.75\% \pm 5.46\%$ & $22.20\% \pm 15.43\%$ \\
CFL & $24.42\% \pm 4.86\%$ & $61.00\% \pm 5.05\%$ & $48.47\% \pm 4.08\%$ & $29.52\% \pm 4.50\%$ & $67.10\% \pm 4.80\%$ & $25.09\% \pm 20.58\%$ \\
\textbf{Ours} & $\textbf{86.08\%} \pm 1.67\%$ & $\textbf{99.07\%} \pm 0.21\%$ & $\textbf{68.13\%} \pm 2.21\%$ & $\textbf{74.52\%} \pm 1.52\%$ & $82.10\% \pm 3.45\%$ & $\textbf{57.09\%} \pm 13.32\%$ \\
\bottomrule
\end{tabular}
\end{table*}


 

\subsection{Datasets}  \label{sec: Dataset}

We evaluated the proposed approach on four well-established datasets: FashionMNIST~\cite{fashionMNIST} \emph{CIFAR10}~\cite{cifar10}, \emph{CIFAR100}~\cite{cifar10}, and \emph{real-data} from the OSCAR corpus~\cite{ortiz-suarez-etal-2020-monolingual} as a motivating real-world application (see Figures~\ref{figure: NLP} and~\ref{figure: pFedLIA Illustration}). Between these datasets, we evaluate on \textbf{two different modalities}, image and text data, and \textbf{two types of heterogeneity}, label space and feature space.

\subsection{Types of Data Heterogeneity} \label{Types of Data Heterogeneity}

The first three datasets (FashionMNIST, CIFAR10, CIFAR100) are used in a supervised classification task. On these datasets, we evaluate our method's performance in a \textbf{label space heterogeneous} setting. Specifically, consider two different scenarios for simulating non-identical data distributions across the clients:
\begin{enumerate}
    \item \textbf{Pathological Non-IID}: In this setting, inspired by related literature (e.g.,~\cite{zhang2020personalized,mcmahan2017communication}, we artificially create five clusters, and partition all the data by assigning each label exclusively to a cluster (e.g., if there are 10 available labels, we split them evenly, 2 in each cluster). For the case of CIFAR100, we group classes in to their superclasses and then assign four super classes per cluster.
    \item \textbf{Noisy Non-IID}: This setting extends the Pathological Non-IID setting by including the possibility of each client having access to an additional randomly selected label. With a probability of 50\%, we assign each client another random label not in their cluster. In CIFAR100, we assign five random labels instead of one. Note that including even more additional random classes for each client results in an ever more IID setting, thus we opted to only add 1.
\end{enumerate}

Meanwhile, the text data from the OSCAR corpus demonstrates an unsupervised learning setting with high \textbf{feature space heterogeneity}. More specifically, we evaluate our approach on a next-word prediction task on Scandinavian languages (Norwegian, Swedish, and Danish). These languages are closely related, and speakers can broadly read all three without great difficulty~\cite{ed7ea7ee31534584be17e2dcd4e7ba7}.

\subsection{Baselines}

We compared to the following baselines: \textbf{(i) FedAvg~\cite{mcmahan2017communication}}: the original Federated Learning aggregation method. A single model trained on all client data. \textbf{(ii) Oracle}: this corresponds to perfect clustering. It can only be used for the Pathological Non-IID setting, where we have clear-cut clusters. This baseline corresponds to the upper bound of any clustering-based personalized FL technique. \textbf{(iii) Local-only}: averaged accuracy over all clients trained only on their local data, without aggregators. \textbf{(iv) FeSEM}~\cite{long2023multi}: a gradient similarity based clustering technique. It is important to note that we provided the correct k (number of clusters) in our simulations, something that \emph{might not be possible in real-world settings.} \textbf{(v) FedFomo}~\cite{zhang2020personalized}: another gradient similarity based technique. \textbf{(vi) CFL}~\cite{ghosh2020efficient}: an iterative clustering strategy that splits clusters into two if the cosine-similarity between weight updates is past a certain threshold. 
\textbf{(vii) FedProto}~\cite{tan2022fedproto}: the clients and server communicate abstract class prototypes used to regularize the training of local models.
We used the FL Bench~\cite{Tan_FL-bench} framework for the classification task, which has most of the aforementioned baselines already implemented. 

\subsection{Evaluation Setup} \label{sec: Evaluation Setup}

Our evaluations are based in a typical Federated Learning scenario. The classification datasets are split amongst the clients, with no data overlap. The dataset is sharded per label into equal sizes and then distributed to the clients, according to the selected heterogeneity method. This data is then further split into a train and validation set, with a ratio of $3:1$. Every simulation has 100 clients, with only 10\% of clients training and communicating gradients per communication round. Our proposed method, pFedLIA, is run after the 20th communication round, in order to first train a warm-up model -- following related methods in the literature, e.g., CFL~\cite{ghosh2020efficient}. Prior to that, the model updates are aggregated via standard FedAvg. After the clustering, we again use FedAvg, but only amongst cluster participants. Recall that our approach can be combined with any FL aggregation method (e.g., FedAvgM~\cite{hsu2019measuring}, or SCAFFOLD~\cite{karimireddy2020scaffold}).

We run each simulation \emph{4 times}, and we report average values and standard deviations. The proposed approach is \textbf{model-agnostic}, \textbf{modality-agnostic}, and can be used with \emph{any} gradient-descent-based ML method.

For the classification tasks we used Res-Net18~\cite{he2016deep}, while for the next-word prediction task we used GPT-2~\cite{radford2019language}. We used pretrained weights for both. Please see the supplementary material for implementation details.

\begin{figure*}[t]
	\centering
	\begin{subfigure}{0.5\linewidth}
	  \centering
	  \includegraphics[width=\linewidth, clip]{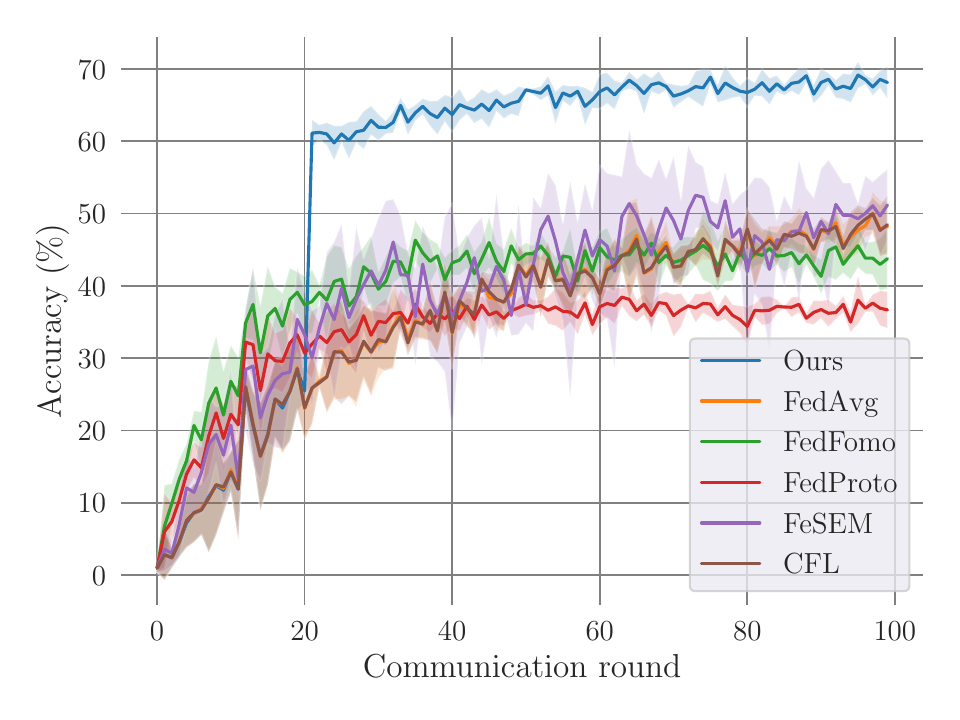}
	  \caption{Pathological Non-IID}
	\end{subfigure}%
	\begin{subfigure}{0.5\linewidth}
	  \centering
	  \includegraphics[width=\linewidth, clip]{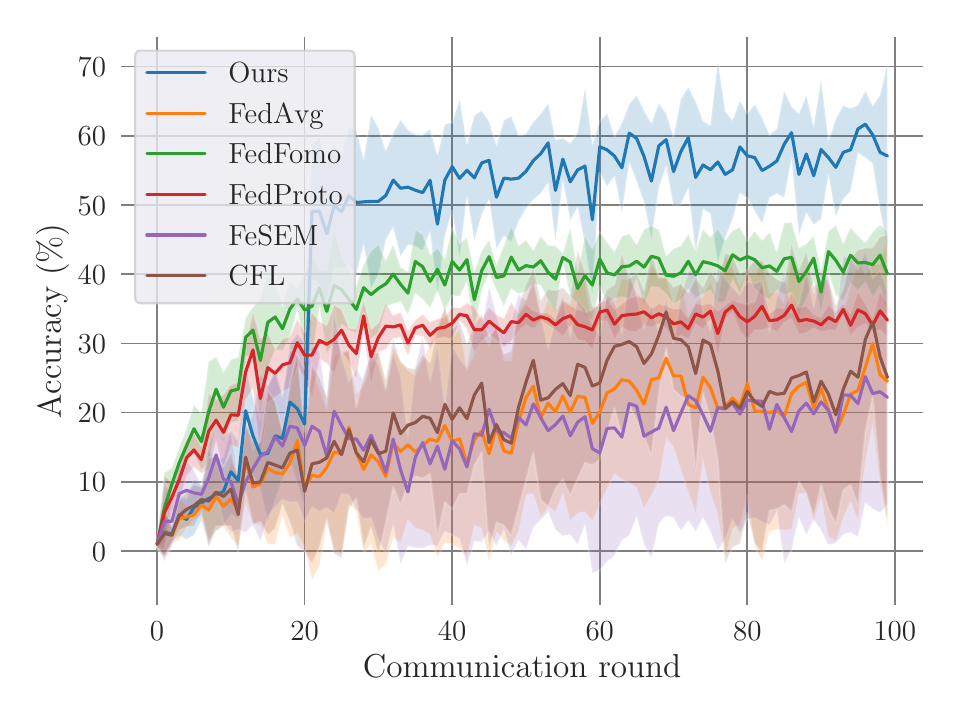}
	  \caption{Noisy Non-IID}
	\end{subfigure}
    
	\caption{Training comparison of our proposed method and all baselines on CIFAR100. Note that we train the `warm-up' model till round 20 (see Section~\ref{sec: Evaluation Setup}), hence the jump for the proposed approach at this time-step.}
	\label{fig: training}
\end{figure*}

\subsection{Results}


Table~\ref{tb: Results} reports comprehensive results on the accuracy of all approaches across all classification datasets, while Figure~\ref{fig: training} depicts the performance across communication rounds.

Our approach \emph{outperforms all baselines} for the pathological setting, across all three datasets, often quite significantly. Importantly, \emph{it matches the Oracle performance}, showcasing that pFedLIA can perfectly cluster clients. 

Compared to the best performing baseline, we outperform FedFomo by 5.1\% and 0.44\% in CIFAR10 and FashionMNIST, respectively, while having \emph{a significant 16.93\% improvement over FeSEM in CIFAR100}. Similarly, in the Noisy Non-IID setting, our approach outperforms the best performing baseline (FedFomo) in CIFAR10 by 4.54\%, and \emph{in CIFAR100 by a significant 17.03\%}. It only achieves slightly lower performance in FashionMNIST, only compared to FedProto (by 5\%) and FedFomo (by 3\%), while outperforming all other baselines. Additionally, our proposed approach achieves significantly faster convergence compared to other techniques, as seen in Figure~\ref{fig: training}. Note that we train the `warm-up' model till round 20 (see Section~\ref{sec: Evaluation Setup}), hence the jump for the proposed approach at this time-step.\footnote{Till round 20, we perform standard FedAvg. Then, within one communication round, we form the clusters and subsequently keep training using FedAvg per cluster.}

Overall, the proposed approach achieves strong performance: \emph{it matches the performance of the Oracle clustering}, and \emph{outperforms the best baseline by up to 17\%.}




\section{Conclusion}


In this work, we propose a clustering-based personalized Federated Learning algorithm that addressed a central challenge in FL: handling the inherent \emph{heterogeneity} in client data distributions. The proposed approach, \emph{pFedLIA}, works in settings with a center, but, importantly, in \emph{distributed peer-to-peer learning settings as well}. Additionally, pFedLIA does \emph{not require a priori knowledge of the number of clusters}. It is both model and modality indifferent. 
pFedLIA utilizes a fast, locally computed Lazy Influence Approximation score to effectively cluster clients in a variety of scenarios, including both label and feature space heterogeneity. We evaluated our approach in a next-word prediction task on the languages, as well as several well-established tasks. \emph{pFedLIA matches the performance of the Oracle clustering, and outperforms the best baseline by up to 17\%.}


\section{Acknowledgments}

P.D. is supported by the European Union’s HORIZON Research and Innovation Programme under grant agreement No. 101120657, project ENFIELD (European Lighthouse to Manifest Trustworthy and Green AI).


\bibliographystyle{plain}
\bibliography{lia_pfl}


\end{document}